# Cancerous Nuclei Detection and Scoring in Breast Cancer Histopathological Images

Pegah Faridi[*,a], Habibollah Danyali[a], Mohammad Sadegh Helfroush[a], and Mojgan Akbarzadeh Jahromi[b]

[a]Department of Electrical and Electronics Engineering, Shiraz University of Technology, Shiraz, Iran
[b]Department of Pathology, School of Medicine, Shiraz University of Medical Sciences, Shiraz, Iran

**Abstract:** Early detection and prognosis of breast cancer are feasible by utilizing histopathological grading of biopsy specimens. This research is focused on detection and grading of nuclear pleomorphism in histopathological images of breast cancer. The proposed method consists of three internal steps. First, unmixing colors of H&E is used in the preprocessing step. Second, nuclei boundaries are extracted incorporating the center of cancerous nuclei which are detected by applying morphological operations and Difference of Gaussian filter on the preprocessed image. Finally, segmented nuclei are scored to accomplish one parameter of the Nottingham grading system for breast cancer. In this approach, the nuclei area, chromatin density, contour regularity, and nucleoli presence, are features for nuclear pleomorphism scoring. Experimental results showed that the proposed algorithm, with an accuracy of 86.6%, made significant advancement in detecting cancerous nuclei compared to existing methods in the related literature.

## 1. INTRODUCTION

Diagnosis and treatment of cancer diseases are one of the most active fields of research. Cancer detection in the primary levels and before metastasis to other organs allows opportunity for successful treatments. According to statistics obtained from the International Agency for Research on Cancer, among the endless list of discovered cancers, breast cancer is one of the most widespread cancers worldwide, particularly among middle-age women [1]. Research has postulated that controlling cancer before spreading through the human body, reduces the mortality rate for cancer patients. Consequently, specialists are investigating solutions for early detection of breast cancer and first stage therapy.

A variety of imaging techniques are employed for cancer detection, diagnosis of cancer metastasis in the body, and cancer degree. Histopathological images are considered for cancer detection and grading which are derived from Aperio XT scanners and Hamamatsu NanoZoomer scanners in subcellular resolution. Because of morphological features of tissue structures, histopathological species are examined by pathologists in a variety of magnification, such as 10X, 20X, and 40X in order to determine the presence of cancer and its level of growth in the human body. Pathologists allocate a degree to each histopathological image in accordance with a defined grading system for each type of cancer. Take, for instance, the Gleason grading system [2] defined for prostate cancer. Bloom and Richardson proposed the first grading system for breast cancer based on structural features found in histopathological images of breast cancer in 1957 [3]. Nottingham modified Bloom and Richardson's grading system, into a system that is currently recognized as a benchmark for diagnosing and grading breast cancer [4]. The Nottingham grading system is based on three conventional criteria; mitotic count, tubule formation, and nuclear pleomorphism. A score from 1 to 3 is assigned to each criterion, with 1 being the lowest and 3 being the highest sign of cancer. The overall grade of the cancer is attained by the sum of the scores of all criteria. In other words, the overall summation of breast cancer scores is in the range of 3 to 9, and then that summation is classified into three grades: 1, 2, or 3.

Although non-invasive methods of imaging such as radiology images are able to detect an abnormal activity in the organs, they cannot explain whether the change is definitely caused by cancer. Consequently, tissue histopathology is amenable for detecting and grading cancers. Histopathological images differ from radiology images because radiology images, such as X-ray, MRI, and ultrasound, are gray level images. The most obvious distinctive feature is the staining process used in slide preparation of biopsy samples, resulting in colorful histopathological images. Histopathological images are large images, approximately around $10^9$ pixels, but radiology images, are composed of only $10^5$ pixels.

*Address correspondence to this author at the Department of Electrical and Electronics Engineering, Shiraz University of Technology, Shiraz, Iran; email: p.faridi@sutech.ac.ir
* present address: Department of Electrical and Computer Engineering, Kansas State University, Manhattan, Ks, US 66506-5204

Diverse computer methods play an important role in evaluating histopathological images. Computer-Aided-Diagnosis (CAD) help physicians achieve more efficient cancer diagnosis and grading. A wide variety of CAD systems are offered for breast cancer detection and grading in histopathological images. These systems are based on either accomplishing one task or all three tasks of the Nottingham grading system in order to present an overall grade of breast cancer [5]–[12].

Among the three criteria of the Nottingham grading system, mitotic count has attracted the most attention among researchers, resulting in a wide range of systems in the literature [13]–[19]. In order to obtain an overall grade for breast cancer, scoring of tubule formation and nuclear pleomorphism is as vital as mitotic count criterion. Unfortunately, however, few works have been proposed regarding those two criteria.

Only a small number of systems have been presented for scoring tubule formation [12], [20]–[23]. Tubule structure contains a white blob in the middle of a continual circle of nuclei. The first step of segmenting tubule structures is to segment the surrounding nuclei; a large number of CAD systems have been proposed for this task [24]–[29].

Nuclear pleomorphism distinguishes the differences between healthy cell nuclei and cancerous cell nuclei. When cancer exists, breast tissue consists of healthy cell nuclei, and deformed cell nuclei which are defined as nuclear pleomorphism. This criterion of the Nottingham grading system concentrates on nuclei shape, size, and the amount of chromatin density. Therefore, nuclei detection and segmentation play a vital role in nuclear pleomorphism grading. Although, a large number of methods have been proposed for nuclei detection and segmentation, the aim of these methods have been to segment healthy nuclei from background which does not yield results if the method must segment cancerous nuclei because the more variations that occur in a cancerous nucleus, the more revealing the distinction is. Therefore, CAD systems proposed for healthy nuclei require alterations in order to be useful for nuclear pleomorphism detection and grading.

In Dalle's work [8], the Gaussian function was employed to design probability distributions of the colors of cell nuclei. Various grades of cancerous cell nuclei possess distinctive mean and variance of the probability function. Each grade of cell nuclei was determined in the proposed system by drawing an analogy between the modeled function and the mean and variance of the probability function. This system was implemented on six samples in which three samples with scores of 2 were detected correctly. However, among the rest of the samples with scores of 3, only one sample was diagnosed accurately, proving that this system cannot perform correctly for cell nuclei segmentation and scoring with scores of 3.

Cosatto et al. [30] proposed a method that uses the active contour model and Support Vector Machine (SVM) classifier. In this work, deformed cell nuclei were distinguished from healthy cell nuclei according to segmented cell nuclei shape, texture and fitness of the their outlines. According to the method results, 92% of segmented cell nuclei were cancerous and 20% of cancerous cell nuclei were not identified. Therefore, in the process of distinguishing cancerous and healthy cell nuclei, a moderate number of cancerous cell nuclei were unaccounted for; if cancerous cell nuclei had scores of 3 but were not identified, grading would be inaccurate.

Dalle et al. [31] proposed a more accurate method based on the assumptions that not all nuclei must be segmented, and that only critical candidates are sufficient for grading. In this method, the region of interest is initially extracted using the method described in [8]. Next, Gamma correction is employed, followed by distance transform and segmentation of a selection of deformed cell nuclei. However, the accuracy of this method decreased due to the increase in chromatin density in the cell nuclei. This method proved to be rarely reliable for diagnosing cell nuclei with scores of 3 because the score increased in relation to the rise of chromatin density.

Although nuclear pleomorphism is an essential criterion for grading breast cancer, it has not been sufficiently studied. Few proposed systems are not entirely reliable because they do not accurately detect and score critical cancerous cell nuclei. Therefore, availability of an accurate method would be extremely beneficial to pathologists.

The proposed method in this paper combines detection, segmentation, and scoring of cancerous cell nuclei in histopathological images of breast cancer. This method is able to accurately detect and score more than 20,000 cancerous cell nuclei without missing any critical nuclei. The proposed CAD system in this paper comprises nuclear pleomorphism criterion in the Nottingham grading system.

The remainder of this paper is organized into descriptive sections. Section 2 includes a detailed description of the proposed system is described in details. Section 3 presents implementation results of the system, and concluding remarks are presented in section 4.

## 2. MATERIALS AND METHOD

### 2.1. Materials

A wide range of materials was employed in this work in order to obtain accurate, reliable results. Possessing the knowledge of several algorithms such as filtering, level set algorithm, Completed Local Binary Pattern (CLBP) features and SVM classifier is the necessity of comprehension of this work.

**2.1.1. Anisotropic diffusion filter** is a helpful tool for image processing because it smoothes the image while preserving as much of the edges as possible. Diffusion across the edges is prevented primarily by a control function. Therefore, this filter removes noise while edges are not smoothed out. Anisotropic diffusion is defined as:

$$\frac{\partial I}{\partial t} = div(c(x, y, t)\nabla I) = \nabla c \cdot \nabla I + c(x, y, t)\Delta I \qquad (1)$$

where *I* is the gray-level image, *t* defines the number of iterations, Δ denotes the laplacian, ∇ denotes the gradient, *div()* is the divergence operator and *c(x,y,t)* is the diffusion coefficient. If this coefficient is constant, then the anisotropic diffusion equation reduces heat function and the filtering would be equivalent to Gaussian blurring. Consequently, *c(x,y,t)* is usually chosen as a function of the image gradient in order to preserve edges in the image and control the rate of diffusion.

**2.1.2. Laplacian of Gaussian (LoG) filter** is a combined filter obtained by Gaussian smoothing and Laplacian filter application. In other words, the LoG filter is a second derivative of the Gaussian function. Edges in the original image manifest themselves as zero-crossings in the Laplacian: One side of an edge of the Laplacian is positive while the other side is negative because an edge produces a maximum or minimum in the first derivative, consequently generating a zero value in the second derivative.

**2.1.3. Difference of Gaussian (DoG) filter** performs edge detection by applying two Gaussian functions on the image, with a unique deviation for each, and subtracting these two functions to yield the result, as shown in equations 4 and 5.

$$DoG(\hat{x}, \sigma) = \frac{1}{2\pi\sigma_1^2} e^{-\frac{\|x^2\|}{2\sigma_1^2}} - \frac{1}{2\pi\sigma_2^2} e^{-\frac{\|x^2\|}{2\sigma_2^2}} \qquad (4)$$

DoG filtering is a convolution of the original image with the difference of two Gaussian filters, defined as follows:

$$G(\hat{x}, \sigma) = \frac{1}{2\pi\sigma_1^2} \int e^{-\frac{\|x-\hat{x}\|^2}{2\sigma_1}} \qquad (5)$$

where *G* is the filtering definition, *DoG* is the equation of the Difference of Gaussian function with two different standard deviations $\sigma_1$ and $\sigma_2$. *I* is the input image and *x* stands for coordinating pixels.

**2.1.4. Level set algorithm** is a set of mathematical methods that attempts to identify points at which the image brightness changes sharply. The level set method is heavily dependent on an initial guess of contour, which is then moved by image driven forces to the boundaries of the desired objects. The initial contour is defined using equation 6, which is based on the level set function with the formulation in equation 7.

$$B = \{(x,y) | \phi(x,y,t) = 0\} \qquad (6)$$

$$\frac{\partial \phi}{\partial t} + F|\nabla \phi| = 0 \qquad (7)$$

In level set algorithm, two types of forces are considered. Internal forces, defined within the curve, are designed to keep the model smooth during the deformation process, while the external forces, computed from the underlying image data, are intended to move the model toward an object boundary.

**2.1.5. Completed Local Binary Pattern (CLBP)** is an improvement of the Local Binary Pattern (LBP) descriptor. An LBP code is computed by comparing a pixel in the center of a region to its neighbors as follows:

$$LBP_{(P,R)} = \sum_{P=0}^{P-1} s(g_P - g_c) 2^P, \; s(x) = \begin{Bmatrix} 1, 0 \leq x \\ 0, x<0 \end{Bmatrix} \qquad (8)$$

where *P* is the total pixels involved in the considered region, *R* is the radius of the neighborhood, $g_c$ is the gray value of the central pixel and $g_p$ is the value of pixels surrounding the central pixel in the neighborhood. Once an LBP pattern is generated for each pixel, a histogram is produced based on LBP patterns such that the original image is transferred to a new space built by a histogram. Construction of the histogram is defined as:

$$H(d) = \sum_{i=1}^{i} \sum_{j=1}^{j} f(LBP_{(P,R)}(i,j), d), d \in [0, D] \qquad (9)$$

where *D* is the maximum of the value of LBP patterns and *f(x,y)* functions as follows:

$$f(x, y) = \begin{Bmatrix} 1, x=y \\ 0, otherwise \end{Bmatrix} \qquad (10)$$

As in equation 8, LBP patterns are computed based on the magnitude of the difference of the gray level of the central pixel and its neighbors. However, in the CLBP descriptor, sign of the difference of the gray levels is also considered along with magnitude of the difference of the gray levels which is considered an improvement for extracting rotation invariant features [32].

**2.1.6. Support Vector Machine (SVM)** is a supervised learning model used for image classification. A SVM training algorithm builds a model that assigns new examples of test members to one category or other categories of the training set. Each member of the training set is marked as belonging to each determined category. SVM can be used with various kernels. Linear SVM is a classifier that separates a set of objects into respective groups by using a line in two-dimensions or so called hyperplane in higher dimensions. The goal of a SVM is to find the optimal separating hyperplane which maximizes the margin of the training data. Given a particular hyperplane, margin can be computed twice as the distance between the hyperplane and the closest data point. In other words, the optimal hyperplane will be the one with the biggest margin so that none of data points of each category could be seen inside the margin.

**2.2. Method**

This research proposes an automatic CAD system for nuclear pleomorphism detection and scoring. Fig. (1) shows a block diagram of the proposed system. The first step of CAD systems is typically preprocessing. One of the major disadvantages with histopathological images, which considerably affect which considerably affect the processing result, is their noisy regions and excessive structures. As a consequence, preprocessing plays a significant role in increasing the accuracy of the histopathological images in a CAD system. Next, objects of interest are segmented through different types of segmentation methods. Segmentation disjoints objects or region of interest from the background so that they can be used in the feature extraction step. Feature extraction transmits information inside images to a smaller and simpler space of features. Finally, every segmented and simpler space of features. Finally, every segmented object is classified according to extracted features. The outputs of classifiers comprise the overall result of the system.

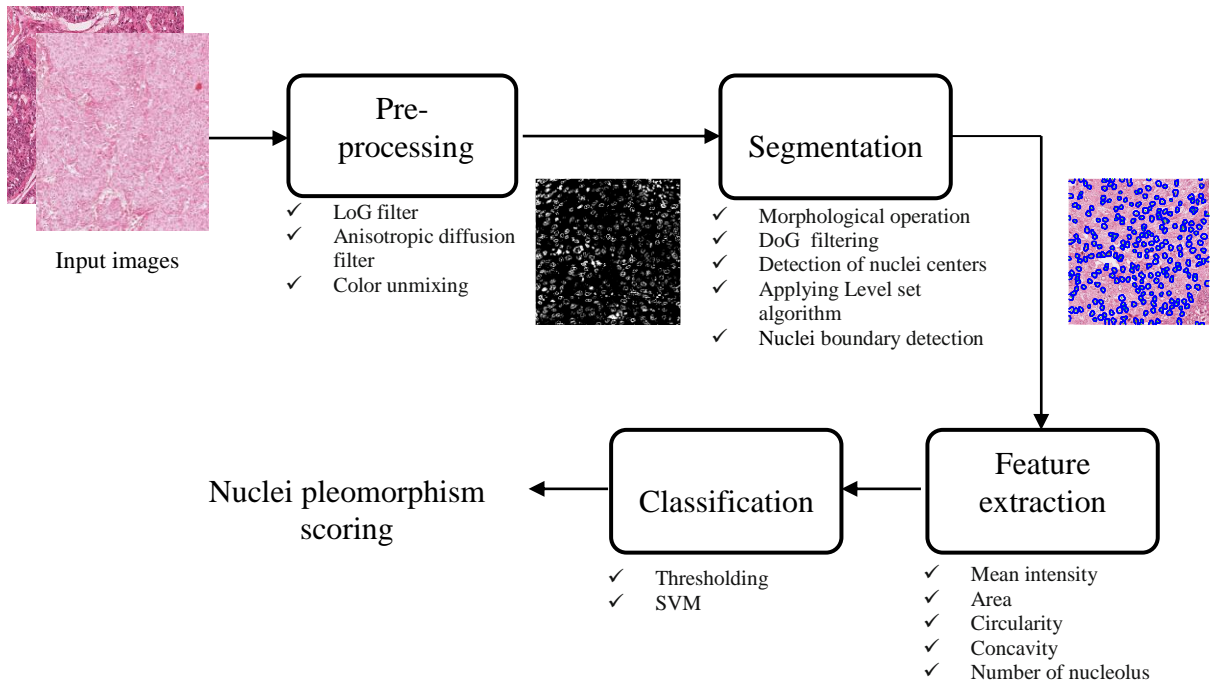

**Fig. (1).** Summarized block diagram of the proposed system

### 2.2.1. Preprocessing

The output image of the preprocessing step should be mostly free of any noise and extra structures, but the main structures should remain unchanged. Therefore, in the proposed system, images are preprocessed using a number of filters and transforms so that redundant structures are removed and only nuclei or structures similar to nuclei remain. The aim of the proposed system is to detect and score nuclei.

Staining biopsy samples is the last step of tissue preparation in order to attain histopathological images. Hematoxylin and Eosin (H&E) are commonly used for staining; H dyes the cell nuclei blue-purple, and E dyes connective tissue and cytoplasm pink. However, in H&E staining, false positive cell nuclei occurs with strong unspecified and noisy background in data that are analysed in the proposed system. Therefore, the preprocessing step becomes essential. Pre-processing operation is performed to the input images as follow:

**Color separation followed by an anisotropic diffusion filter:** Histopathological images may appear in different colors according to the amount of H&E used in the staining process. As mentioned, cell nuclei can be demonstrated in blue or purple; however, cell nuclei are not easily separable from other structures in the image when cell nuclei are shown in purple. Therefore, an unmixing process for separating H&E channels is necessary. Because the proposed system tends to segment cell nuclei, all subsequent approaches are accomplished on the gray level image of the H channel which is gained by separating H&E stains [34]. Next, an anisotropic diffusion filter is applied to the gray level image of the H channel in order to smooth and reduce image noise without removing significant parts of the image content, typically edges, leading to a higher contrast between background and edges.

Then, as shown in Fig. (**2**), edges are extracted using LoG filtering. Because both applied filters aim to alleviate redundant structures and preserve edges, irrelevant elements and noise are rarely noticed in the binary output image.

### 2.2.2. Nuclear pleomorphism segmentation

In the proposed system, the segmentation step consists of two levels. As illustrated in Fig. (**3**), in the first level, the center of nuclei are detected using several morphological operations, and DoG filter. Second, nuclei boundaries are segmented by the level set algorithm. In other words, nuclei are detached from the background and other structures.

• **Morphological operation**: Cancer not only modifies the shape of cell nuclei, but it also changes existing structures in the tissue so that large clusters of cell nuclei are constructed. These clusters are considered for nuclear pleomorphism scoring. Morphological operations are applied to the output binary image of the preprocessing step in order to generate and maintain connected clusters of objects and remove isolated objects. One iteration closing, filling, and erosion are the employed morphological operations, respectively. The input image for the first level of the segmentation step

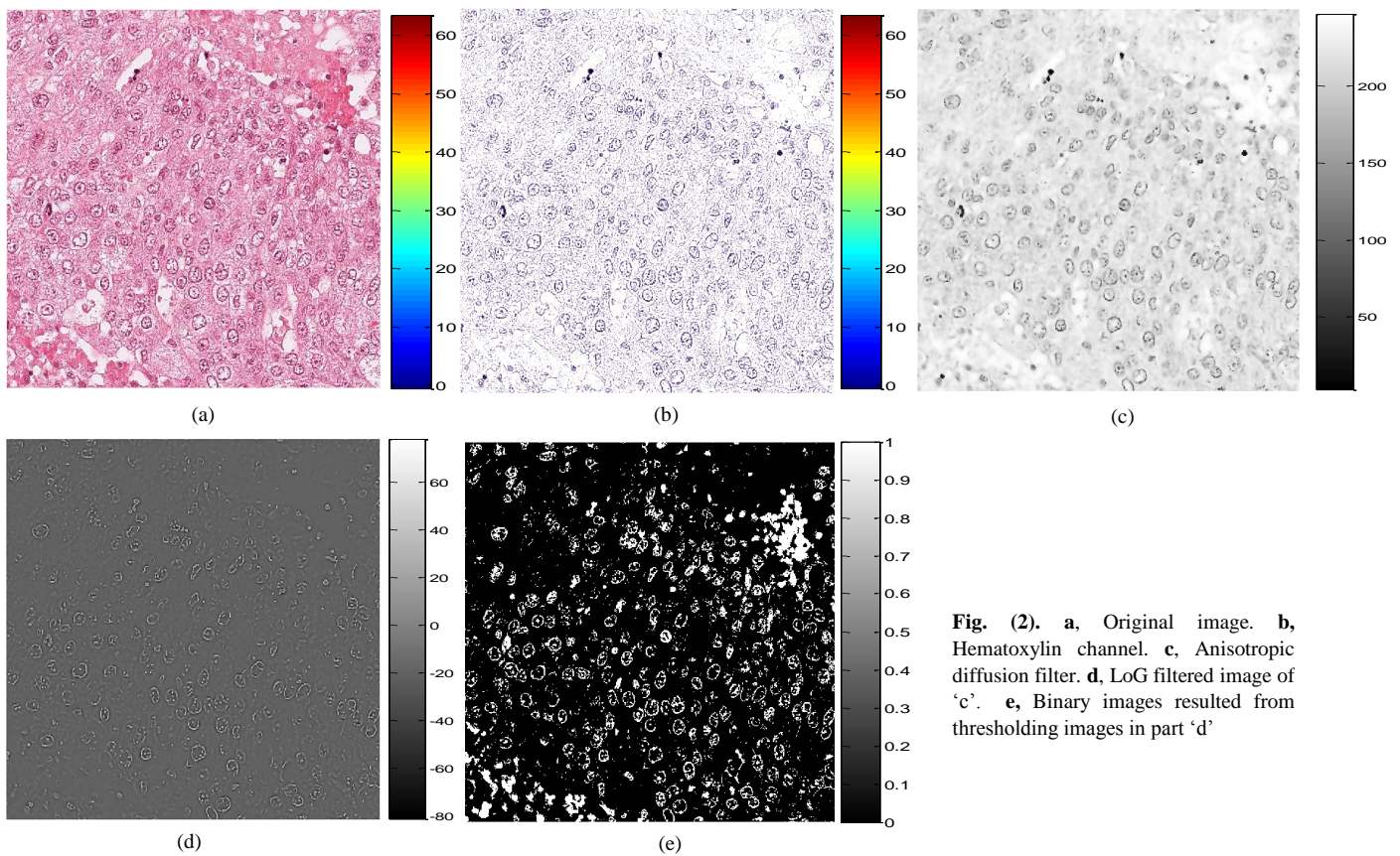

**Fig. (2). a**, Original image. **b**, Hematoxylin channel. **c**, Anisotropic diffusion filter. **d**, LoG filtered image of 'c'. **e**, Binary images resulted from thresholding images in part 'd'

includes a wide range of disjointed objects obtained from the thresholded image of the preprocessing step. In order to achieve structures similar to real cancerous nuclei in histopathological images, cell nuclei should appear in joint clusters and isolated objects should remain isolated. This issue can be addressed by the closing operation, which contributes to connected objects as boundaries followed by the filling operation, which fills the blobs that appeared after closing operation. As a result of these operations, a large number of clusters that contain cell nuclei are revealed and isolated objects remain unchanged.

Although the morphological operations create a binary image similar to the actual histopathological image, a cancerous nuclei centroid connot be detected unless isolated objects are eliminated, thereby requiring the erosion operation with the structural element size of healthy cell nuclei in our images which our collaborated pathologist has recognized them as the healthy ones so the area and radius of them have been measured with this proposed method. Fig. (**4**)(**a**) shows that this operation eliminates isolated objects and decreases cluster thicknesses, resulting in getting closer to the center of each nuclei.

• **DoG filtering:** The binary image resulted from morphological operations is filtered by DoG filtering after which distinct blobs inside nuclei emerge by thresholding the filtered image. In other words, after thresholding the filtered image, the blobs no longer represent the clusters because clusters do not exist anymore and have converted to disjointed blobs. Blobs are bright on dark or dark on bright.

In this method, blobs are bright on dark which can be detected by DoG filtering. This filter performs edge detection, which works by applying two different Gaussian function on the image, with a different blurring radius for each function, and subtracting them to yield the result. The most important parameters are the blurring radii for the two Gaussian functions. Increasing the smaller radius tends to give thicker-appearing edge and decreasing the larger radius tends to increase the threshold for recognizing objects as edge. Therefore, radii should be selected regarding the expected result of the filtering process. In this proposed method, standard deviations of 4 and 10 was chosen empirically as they lead to the best result and they were constant for all the images in the database.

Regarding the thresholding, because the expected result of thresholding the filtered image is achieving the nuclei centroid, the threshold value has to be large enough to keep the bright points which, according to the Fig. (4). b, are the centers of blobs. In this proposed method, the optimum threshold is to keep all the pixels with values above 200 in images with the range of [0 255], as it's illustrated in Fig. (4), b.

• **Detection of nuclei centers**: Center of each blob is nuclei centroid. Fig. (**5**) shows a sample of identified centroids by utilizing the proposed system. This figure illustrates that there are some centroid selections in which two centroids exist within the same nucleus. This does not have any influence on the CAD system performance because it's solved in the first part of the second level of segmentation that is constructing an initial contour. When the center of deformed nuclei are detected, second level of segmentation step begins. Nuclei boundaries are extracted using level set method; an edge detector which its' efficiency peripherals to both initial contour and input image.

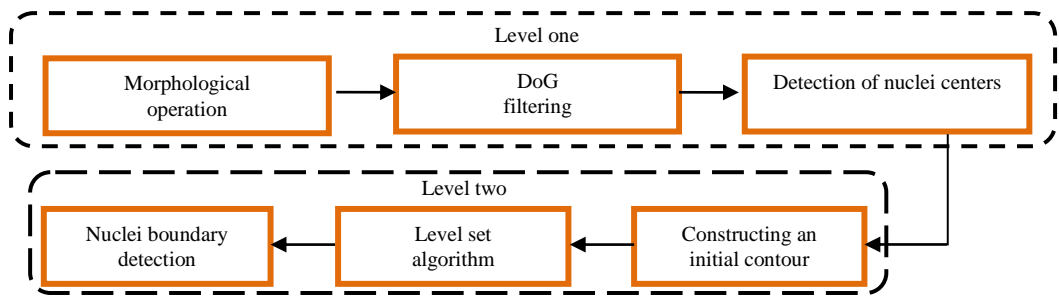

**Fig. (3).** Internal block diagram of the segmentation step of the proposed system

- **Constructing an initial contour:** In the proposed system, initial contour is constructed by the use of the detected center of cancerous nuclei. First, detected centers are dilated and multiplied by the output image of preprocessing step which consists of a minimum number of redundant objects. Because of multiplication operation, excess thicknesses of boundaries that appear during the dilation process and wrongly detected centers are diminished. Moreover, closing and filling operations are applied to the product result to create closed contours. The final initial contour is shown in Fig. (**6**), **b**.

- **Applying level set algorithm:** Distance Regularized Level Set Evolution (DRLSE) [33] is applied on the constructed initial contours, while its internal input image is altered, because the input image influences the edge detection function in the DRLSE algorithm. Therefore, input image should be employed such that it gives the edge detection function the ability to distinguish nuclei boundaries from the background. In order to change the input image to achieve this goal, anisotropic diffusion filter is applied on the gray level of the original image. Filtered image is considered as an input for DRLSE method. This filtered input image not only affects the result of DRLSE method but also, lowers the inaccuracy of the cancerous nuclei segmentation caused by some selections of nuclei centroids. As demonstrated in Fig. (**5**), there are some centroid selections that are in-between nuclei. In this case, the nucleus which has a higher gradient in the input image of level set algorithm is segmented. The other nucleus which is eliminated does affect the accuracy of segmentation but does not influence the nuclei scoring because in our method, even if one nucleus with score 2 or 3 in a 40X image is segmented, the general score of the 20X image would be 2 or 3, respectively. Since in a 20X image with score 2 or 3, a large majority of nuclei have the same score, losing just few of them during the segmentation does not modify the general score and grading. When initial contour and input image are acquired as explained above, iterations begin. Each nucleus iterates separately. In other words, number of applying DRLSE method equals to number of detected nuclei in the first part of segmentation step. Consequently, final contours which are the boundaries of all nuclei are acquired. Indeed, nuclei are distinguished form background, and are ready for feature extraction and nuclear pleomorphism scoring.

### 2.2.3. Feature extraction

Effective and adequate features must be extracted from segmented nuclei in order to gain an accurate scoring. In this work, various features are extracted in proportion to the criteria that influence nuclear pleomorphism scoring. These criteria involve the variation of nuclei size in the population, so called anisonucleosis, chromatin density, contour regularity, and number of revealed nucleoli which are suggested by ICPR 2014 contest in the context of nuclear atypia scoring in breast cancer histopathological images [35]. Decent features should be extracted for each criterion. Decent features are the ones that have the ability of illustrating every segmented nucleus score with respect to the four criteria.

- **Area** is one of the essential features extracted for scoring in terms of anisonucleosis criterion. Area of every single nucleus is computed.

- **Mean intensity** is the second feature which is utilized for measuring the chromatin density. The uniformity of chromatin in healthy nuclei disappears, and becomes dense gradually. The greater is the score of deformed nuclei, the denser the chromatin is. Therefore, gray level inside healthy nuclei differs from various cancerous nuclei which enables the system to use the mean intensity feature for scoring nuclear pleomorphism on account of chromatin density criterion.

- **Circularity** is morphological feature utilized for nuclear pleomorphism scoring on the basis of contour regularity criterion. Healthy nuclei are circular, but cancerous nuclei may not have circular contour. Their contour may have become angled and irregular. To solve this irregularity problem, circularity feature is used. Circularity is calculated using the equation 11 in which P stands for perimeter and A is regarded as the area of the object.

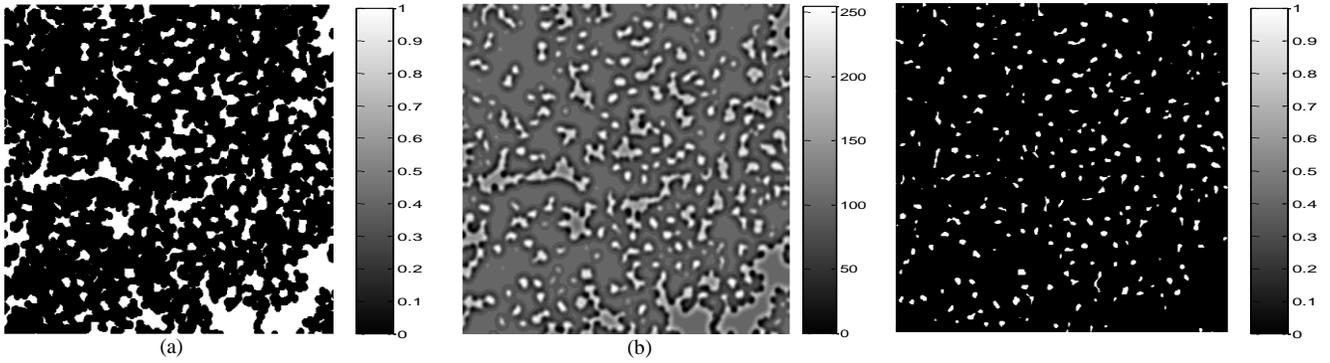

**Fig.(4). a**, Applying morphological operation on the output binary image of preprocessing step. **b**, DoG filtering. **c**, Thresholding filtered image.

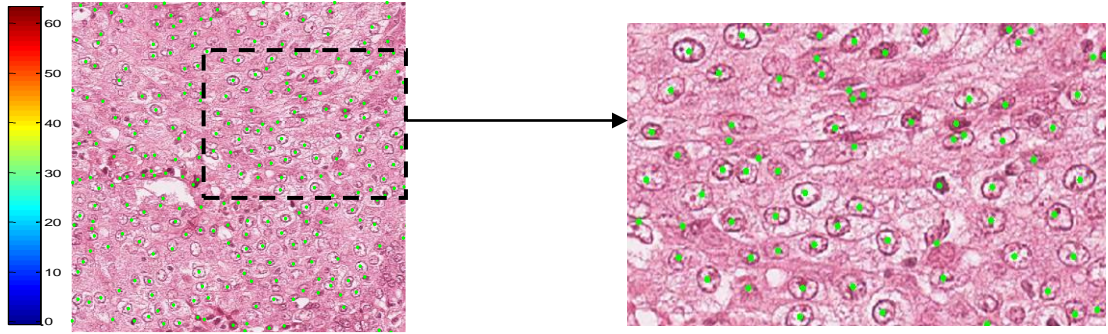

**Fig.(5).** A sample of center of cell nuclei detected by the proposed method

$$Circularity = \frac{P^2}{A} \qquad (11)$$

- **Nucleoli** are an integral part of nuclei. Although nucleoli exist in all healthy nuclei, they are concealed due to uniform chromatin. As a result, nucleoli would not be revealed unless chromatin becomes dense, which occurs in cancerous nuclei, particularly nuclei with score 2 and 3. Therefore, the number of unfolded nucleoli among the nuclei population can be a criterion for nuclear pleomorphism scoring. The Block diagram of this method is shown in Fig. (**7**) which in turn is a subset of feature extraction block in the major proposed CAD system represented in Fig. (**1**). As for nucleoli detection, train and test systems are typically employed. Training and testing samples are segmented using bilateral filtering followed by applying a gamma correction function on blue channel of filtered image. The aim of applying morphological opening operation is to delete points so that the non-nucleoli points are removed, as nucleoli dimension obviously are more than one pixel. In addition, circularity feature is used to divide the detected candidates into two groups, circular and non-circular candidates. Since nucleoli have circular contour, the detected group with circular candidates are selective nucleoli test samples. Training set is selected in one special image the same as method shown in Fig. (**7**), and texture features are extracted using CLBP algorithm. Our collaborated pathologist has identified the nucleoli and non-nucleoli candidates among the segmented training set which gives the proposed method the ability of recognizing the nucleoli and non-nucleoli candidates through the extracted CLBP features. In other words, the training set containing two distinct groups of nucleoli and non-nucleoli candidates with known CLBP features are numbered group 1 and 2, respectively, are inputs to SVM classifier. Next, the selective nucleoli test samples segmented in each image is divided into two groups named 1 or 2 using SVM classifier with linear kernel, which number 1 refers to nucleoli group of test samples while group 2 contains the non-nucleoli ones.

### 2.2.4. Classification

Nuclear pleomorphism scoring is performed using four criteria mentioned in section 2.2.2. The segmented cancerous nuclei are scored on the basis of the difference among features extracted for cancerous nuclei and the same features extracted for healthy nuclei that is completely delineated in Table 1.

As mentioned in section 2.2.2., our collaborated pathologist has recognized healthy cell nuclei in the images so they are segmented with this proposed method and all the four features; area, mean intensity, circularity and concavity and nucleoli, are computed and considered as the normal features which are required for scoring the cancerous cell nuclei based on Table 1.

## 3. RESULTS AND DISCUSSIONS

Histopathological images of breast cancer employed in this proposed method have been derived from International Conference on Pattern Recognition (ICPR, 2014) contest of detection of mitosis and evaluation of nuclear atypia score in breast cancer histological images [35]. The slides are stained with standard H&E dyes and they have been scanned by Aperio XT scanners with a scale of 0.2456 μm per pixel and magnified at 20X and 40X magnification [35]. 34 images with 20X magnification are examined in this test, in which they consist of 4 frames scored 1, 20 frames scored 2 and 10 frames scored 3 according to the ground truth. 20X images are used for nuclear pleomorphism scoring. On the other hand, scores given for each feature by collaborating pathologists in the mentioned contest are associated to 40X images, and only the general score is allocated to 40X images. Therefore, the proposed system is applied on each quarter of each 20X images so that the system is able to use the scores for each feature. In other words, each 20X image is divided into four sections that on each of them, all steps of proposed method are applied because each quarter of a 20X image is the same as a 40X image, and they are different only in two tasks. One difference is the smaller spatial dimension of the quarter image which leads to a higher pace of the system. Secondly, details and redundant structures are found more in a 40X image than in a quarter of 20X one which may cause errors in the detection and segmentation steps.

Since the ground truth of the contest only contains scores allocated to each feature and an overall score for each 20X image, an alternative ground truth is provided by the association of an expert pathologist, who collaborated with us in this project, in order to increase the accuracy of the proposed CAD system. In the second ground truth, all nuclei in all 20X images are marked manually by the collaborated pathologist. As a result, center of detected nuclei can be compared to this second ground truth provided by the cooperation of our team and an expert pathologist leading to a reliable outcome and comparison. Therefore, the accuracy of the results provided by the proposed system also is more reliable compare to methods exist in the literature.

Moreover, nuclei scoring with a high precision require an accurate segmentation of nuclei boundaries. Consequently, the segmented boundary of each nucleus is examined by the collaborated pathologist so that it was determined whether the segmented boundary is accurate and decent for scoring the nucleus or not.

The average accuracy of detected centers and segmented boundaries are demonstrated in Table 2. In [31], which is one of the most relevant and known methods in this regard, the error average of extracted boundaries of all nuclei, and nuclei with score 2 and 3 are presented. Table 3, draws a parallel between the accuracy error offered by [31] and the accuracy proposed in this paper.

Table 4, draws an analogy between scores acquired by the proposed system and sores which exist in the ground truth of contest. The accuracy is estimated by formula in equation 12.

$$Accuracy(ACC) = \frac{T_P + T_N}{T_P + T_N + F_P + F_N} \quad (12)$$

Abbreviations of $T_P$, $F_P$, $T_N$, $F_N$ in equation 12 are:

**True Positive ($T_P$)**: correctly identified nuclei and their boundaries.
**False Positive (FP)**: incorrectly identified non nuclei and their boundaries.

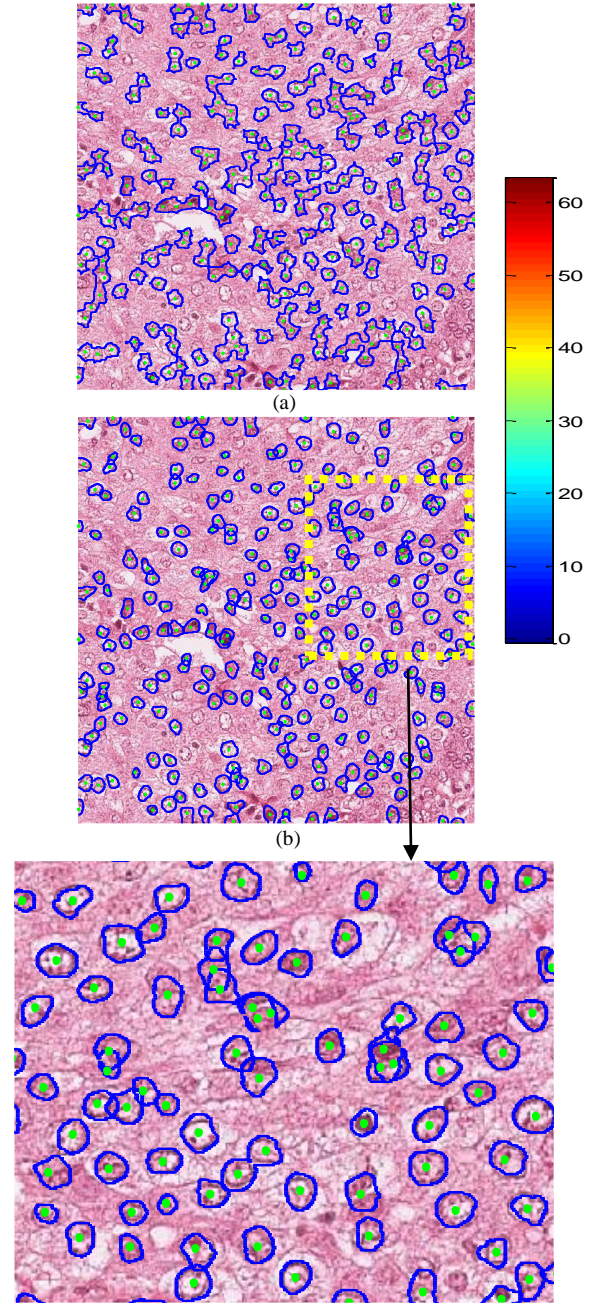

**Fig.6. a**, Initial contour. **b**, Boundary segmented by level set method

**True Negative ($T_N$)**: correctly rejected non nuclei and their boundaries.

**False Negative ($F_N$)**: incorrectly rejected nuclei and their boundaries.

The results of the scores acquired by the proposed system are compared to the scoring accuracy on account of the same features obtained in [30] using the F-measure introduced in equation 13. As it is visualized in Table 5, the scoring accuracy has dramatically increased by the proposed method. In the comparison of the method proposed in this work with the method presented in [31], which was also applied to the dataset used in this paper, the number of revealed nucleoli criterion has not been considered as it is not presented in anyother published proposed methods in this regard. According to the score of every feature, a general score is dedicated to each quarter image, and the whole slide score, in fact the score of nuclear pleomorphism element of the Nottingham Grading System for a 20X magnified image is computed as illustrated in Table 6.

$$F - measure = \frac{2 \times precision \times recall}{precision + recall} \quad (13)$$

$$recall = \frac{T_P}{T_P + T_N} \quad (14)$$

$$precision = \frac{T_P}{T_P + F_P} \quad (15)$$

Final results are summarized in a confusion matrix shown in Table 7. Among the 34 examined images, all the scored 1 images are properly scored. Two images with score 2 and one image with score 3 are designated wrong according to the ground truth offered by collaborating pathologists in the contest.

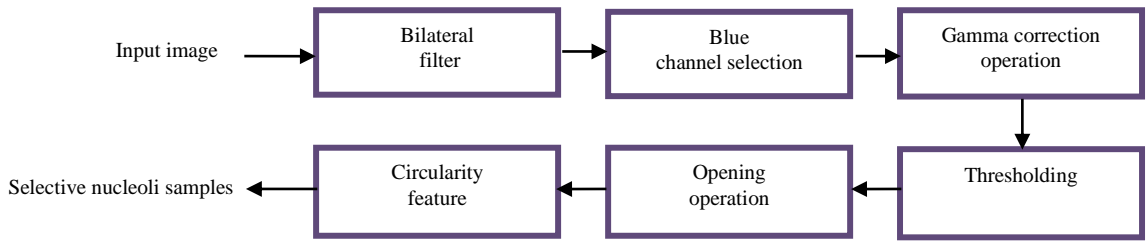

**Fig.(7).** Flowchart of nucleoli detection

**Table 1. Criteria for scoring presented in contest ICPR 2014 [35]**

| Criterion | Score | Description |
| --- | --- | --- |
| | 1 | 0 to 30% of tumour cells have nucleoli size bigger than nucleoli of normal cells |
| Size of nucleoli | 2 | 30 to 60% of tumour cells have nucleoli size bigger than nucleoli of normal cells |
| | 3 | more than 60% of tumour cells have nucleoli size bigger than nucleoli of normal cells |
| | 1 | 0 to 30% of tumour cells have chromatin density higher than normal cells |
| Density of chromatin | 2 | 30 to 60% of tumour cells have chromatin density higher than normal cells |
| | 3 | more than 60% of tumour cells have chromatin density higher than normal cells |
| | 1 | 0 to 30% of tumour cells have nuclear contour more irregular than normal cells |
| Regularity of nuclear contour | 2 | 30 to 60% of tumour cells have nuclear contour more irregular than normal cells |
| | 3 | more than 60% of tumour cells have nuclear contour more irregular than normal cells |
| | 1 | Within the population of tumour cells, all nuclei are regular and/or nuclei size is not bigger than twice the size of normal epithelial cell nuclei |
| Anisonucleosis | 2 | for cases that are not fitting neither with case 1 nor with case 3 |
| (Size variation within a population of nuclei) | 3 | Within the population of tumour cells, either nuclei size are irregular or nuclei size is bigger than 3 times the size of normal epithelial cell nuclei |

**Table 2. Accuracy of the proposed method's simulation result in nuclei detection and boundary segmentation**

| Simulation result | Center of nuclei | Nuclei boundaries |
|---|---|---|
| Maximum accuracy | 92% | 93% |
| Average accuracy | 85.7% | 86.6% |

**Table 3. Comparison between method proposed in this work with Dalle method [31]**

| Evaluation results | Method proposed in this paper | Dalle method [31] |
|---|---|---|
| Overall accuracy error | 13.4% | 15.9% |
| Accuracy error of frames scored 2 | 13% | 15.3% |
| Accuracy error of frames scored 3 | 13.3% | 16.5% |

**Table 4. Scoring accuracy in a quarter of 20X images**

| Criteria | Scoring ACC |
|---|---|
| Anisonucleosis | 82% |
| Nucleoli | 75% |
| Chromatin density | 93.75% |
| Contour regularity | 100% |
| Average | 87.7% |

**Table 5. Comparison of F-measure for method proposed in this work with method presented in [30]**

| Features | F-measure for Proposed method | F-measure for Method used in [30] |
|---|---|---|
| Size of nuclei | 87% | 73% |
| Contour regularity | 100% | 73% |
| Chromatin density | 96.96% | 71% |

**Table 6. Score determination mode for a 20X image**

| Score of a 20X magnified image | How to set up the score |
|---|---|
| 1 | If the score of each quarter equals to 1. |
| 2 | If the score of at least on quarter equals to 2. |
| 3 | If the score of at least on quarter equals to 3. |

**Table 7. Confusion matrix of final results of the proposed system**

| Offered scores in the contest [35] \ Allocated scores by the proposed systems | 1 | 2 | 3 |
|---|---|---|---|
| 1 | 4 | 0 | 0 |
| 2 | 0 | 18 | 2 |
| 3 | 0 | 1 | 9 |

## CONCLUSION

An automatic CAD system is proposed for nuclear pleomorphism detection, segmentation, and scoring. In this system, noisy regions and redundant structures are removed by using effective filters and transforms in the preprocessing step so that the accuracy of detected nuclei increases which sets stage for accurate boundary segmentation. Cancerous nuclei are detected with an average accuracy of 85.7%. Next, boundaries of detected nuclei are segmented with 86.4% accuracy which ends in the high precision of nuclear pleomorphism scoring since cell nuclei scoring is strongly associated to the accuracy of segmented nuclei boundaries. The proposed system is finalized by scoring segmented nuclei according to the extracted features on the basis of four criteria that draw distinction between cancerous nuclei and healthy nuclei. In 34 images examined in this system which are derived from an international contest on the evaluation of nuclear atypia score, ICPR, 2014, only 4 images are misdiagnosed which is a clear cut visualization of an accurate system.

# REFERENCES

[1] Release P. Latest world cancer statistics Global cancer burden rises to 14 . 1 million new cases in 2012 : Marked increase in breast cancers must be addressed. Int Agency Res Cancer, World Heal Organ 2013:2012–4.

[2] Humphrey PA. Gleason grading and prognostic factors in carcinoma of the prostate. Mod Pathol 2004;17:292–306.

[3] Bloom HJG, Richardson WW. Histological grading and prognosis in breast cancer: a study of 1409 cases of which 359 have been followed for 15 years. Br J Cancer 1957;11:359.

[4] Wootton R, Springall DR. Image analysis in histology: Conventional and confocal microscopy. CUP Archive; 1995.

[5] Jeong H-J, Kim T-Y, Hwang H-G, Choi H-J, Park H-S, Choi H-K. Comparison of thresholding methods for breast tumor cell segmentation, IEEE; 2005, p. 392–5.

[6] Dundar MM, Badve S, Bilgin G, Raykar V, Jain R, Sertel O, et al. Computerized classification of intraductal breast lesions using histopathological images. IEEE Trans Biomed Eng 2011;58:1977–84. doi:10.1109/TBME.2011.2110648.

[7] Brook A, El-Yaniv R, Issler E, Kimmel R, Meir R, Peleg D. Breast cancer diagnosis from biopsy images using generic features and SVMs 2007.

[8] Dalle J-R, Leow WK, Racoceanu D, Tutac AE, Putti TC. Automatic breast cancer grading of histopathological images, IEEE; 2008, p. 3052–5.

[9] Singh S, Gupta PR. Breast cancer detection and classification using neural network. Int J Adv Eng Sci Technol 2011;6.

[10] Doyle S, Agner S, Madabhushi A, Feldman M, Tomaszewski J. Automated grading of breast cancer histopathology using spectral clustering with textural and architectural image features, IEEE; 2008, p. 496–9.

[11] Tay C, Mukundan R, Racoceanu D. Multifractal analysis of histopathological tissue images 2011.

[12] Naik S, Doyle S, Agner S, Madabhushi A, Feldman M, Tomaszewski J. Automated gland and nuclei segmentation for grading of prostate and breast cancer histopathology, IEEE; 2008, p. 284–7.

[13] Veta M, van Diest PJ, Willems SM, Wang H, Madabhushi A, Cruz-Roa A, et al. Assessment of algorithms for mitosis detection in breast cancer histopathology images. Med Image Anal 2015;20:237–48. doi:10.1016/j.media.2014.11.010.

[14] Tashk A, Helfroush MS, Danyali H, Akbarzadeh-jahromi M. Automatic detection of breast cancer mitotic cells based on the combination of textural, statistical and innovative mathematical features. Appl Math Model 2015. doi:10.1016/j.apm.2015.01.051.

[15] Wang H, Cruz-roa A, Basavanhally A, Gilmore H, Shih N, Feldman M, et al. Mitosis Detection in Breast Cancer Pathology Images by Combining Handcrafted and Convolutional Neural Network Features n.d.:1–13.

[16] Irshad H, Gouaillard a, Roux L, Racoceanu D. Multispectral Spatial Characterization : Application to Mitosis Detection in Breast Cancer Histopathology. Computerized Medical Imaging and Graphics 2014; 38:390-402.

[17] Tashk A, Helfroush MS, Danyali H, Akbarzadeh M. An automatic mitosis detection method for breast cancer histopathology slide images based on objective and pixel-wise textural features classification. 5th Conf. Inf. Knowl. Technol., IEEE; 2013, p. 406–10. doi:10.1109/IKT.2013.6620101.

[18] Veta M., van Diestb P., Pluim J. Detecting mitotic figures in breast cancer histopathology images. in Proc. of SPIE Medical Imaging, 2013.

[19] Khan A.M., Eldaly H., Rajpoot N.M. A gamma-gaussian mixture model for detection of mitotic cells in Breast Cancer histopathology images. J. Pathol.Inf. 1 (11) (2013) 11 pages.

[20] Maqlin P, Thamburaj R, Mammen JJ, Nagar AK. Automatic Detection of Tubules in Breast Histopathological Images, Springer; 2013, p. 311–21.

[21] Tutac AE, Racoceanu D, Putti T, Xiong W, Leow W-K, Cretu V. Knowledge-guided semantic indexing of breast cancer histopathology images. vol. 2, IEEE; 2008, p. 107–12.

[22] Basavanhally A, Yu E, Xu J, Ganesan S, Feldman M, Tomaszewski J, et al. Incorporating domain knowledge for tubule detection in breast histopathology using O'Callaghan neighborhoods. vol. 7963, International Society for Optics and Photonics; 2011, p. 796310.

[23] A. Nguyen, et. al., 2015, Automatic glandular and tubule region segmentation in histological grading of breast cancer, Proc. SPIE9420, Medical Imaging, 2015: Digital Pathology.

[24] Petushi S, Katsinis C, Coward C, Garcia F, Tozeren A. Automated identification of microstructures on histology slides, IEEE; 2004, p. 424–7.

[25] Veta M, Huisman A, Viergever MA, van Diest PJ, Pluim JP. Marker-controlled watershed segmentation of nuclei in H&E stained breast cancer biopsy images, IEEE; 2011, p. 618–21.

[26] Veta M, van Diest PJ, Kornegoor R, Huisman A, Viergever MA, Pluim JP. Automatic Nuclei Segmentation in H&E Stained Breast Cancer Histopathology Images. PLoS One 2013;8:e70221.

[27] A. Veillard, et al., 2013, Cell nuclei extraction from breast cancer histopathology images using colour, texture, scale and shape information, Diagnostic Pathology.

[28] Wienert S, Heim D, Saeger K, Stenzinger A, Beil M, Hufnagl P, et al. Detection and segmentation of cell nuclei in virtual microscopy images: a minimum-model approach. Sci Rep 2012;2.

[29] Al-Kofahi Y, Lassoued W, Lee W, Roysam B. Improved automatic detection and segmentation of cell nuclei in histopathology images. Biomed Eng IEEE Trans 2010;57(4):841–52.

[30] Cosatto E, Miller M, Graf HP, Meyer JS. Grading nuclear pleomorphism on histological micrographs, IEEE; 2008, p. 1–4.

[31] Dalle J-R, Li H, Huang C-H, Leow WK, Racoceanu D, Putti TC. Nuclear pleomorphism scoring by selective cell nuclei detection. IEEE Workshop on Applications of Computer Vision; 2009 Dec 7-8; Snowbird, Utah.

[32] Z. Guo, L. Zhang, and D. Zhang, "A completed modeling of local binary pattern operator for texture classification," IEEE Trans. Image Process., vol. 19, no. 6, pp. 1657–1663, Jun. 2010.

[33] C.Li,C.Xu,C.Gui,andM.D.Fox,"Distance regularized level set evolution and its application to image segmentation," IEEE Trans. Image Process., vol. 19, no. 12, pp. 3243–3254, Dec. 2010.

[34] Tadrous PJ. Digital stain separation for histological images. J Microsc 2010;240:164–72.

[35] Dataset for nuclear atypia scoring in breast cancer histological images: An ICPR 2014 contest. Available from:

http://ipal.cnrs.fr/data/mitosis_atypia_2014/icpr2014_Mitosis_Atipya_DataDescription.pdf
last visit: Sep 7, 2014.